\documentclass[journal,twoside,web]{ieeecolor}

\usepackage{generic}
\usepackage{cite}
\usepackage{amsmath,amssymb,amsfonts}
\usepackage{algorithmic}
\usepackage{graphicx}
\usepackage{algorithm,algorithmic}
\usepackage{hyperref}
\hypersetup{hidelinks=true}
\usepackage{textcomp}
\usepackage{booktabs}
\usepackage{soul}
\usepackage{subcaption}
\usepackage{paralist}
\usepackage{flushend}
\usepackage{colortbl}
\usepackage{tikz}
\usepackage{multirow}
\usepackage{float}
\usepackage{url}
\usepackage{adjustbox}
\usepackage{lipsum}
\usepackage{tabularx,ragged2e,booktabs}
\usetikzlibrary{positioning, arrows.meta, shapes.geometric, fit}
\usetikzlibrary{trees}
\usepackage{tcolorbox}
\usepackage{array}
\usepackage{pgfplots}
\usepackage{pgf-pie}
\usepackage{caption}
\usepackage{xcolor}
\pgfplotsset{compat=1.18}

\newcommand{\papertitle}{DENSE: Longitudinal Progress Note Generation with Temporal Modeling of Heterogeneous Clinical Notes Across Hospital Visits}

\def\BibTeX{{\rm B\kern-.05em{\sc i\kern-.025em b}\kern-.08em
    T\kern-.1667em\lower.7ex\hbox{E}\kern-.125emX}}
\begin{document}

\title{\papertitle}

\author{Garapati Keerthana, Manik Gupta
\thanks{Garapati Keerthana and Manik Gupta are with the Department of Computer Science and Information Systems, Birla Institute of Technology and Science, Pilani, Hyderabad Campus, Jawahar Nagar, Kapra Mandal, Medchal District, Telangana 500078, India (e-mail: p20240505@hyderabad.bits-pilani.ac.in; manik@hyderabad.bits-pilani.ac.in). ORCID: 0009-0009-5116-1553 (G.K.); 0000-0002-4977-4299 (M.G.).}%
}

\maketitle

\begin{abstract}
Progress notes are among the most clinically meaningful artifacts in an Electronic Health Record (EHR), offering temporally grounded insights into a patient's evolving condition, treatments, and care decisions. Despite their importance, they are severely underrepresented in large-scale EHR datasets. For instance, in the widely used Medical Information Mart for Intensive Care III (MIMIC-III) dataset, only about $8.56\%$ of hospital visits include progress notes, leaving gaps in longitudinal patient narratives. In contrast, the dataset contains a diverse array of other note types, each capturing different aspects of care.

We present DENSE (Documenting Evolving Progress Notes from Scattered Evidence), a system designed to align with clinical documentation workflows by simulating how physicians reference past encounters while drafting progress notes. The system introduces a fine-grained note categorization and a temporal alignment mechanism that organizes heterogeneous notes across visits into structured, chronological inputs. At its core, DENSE leverages a clinically informed retrieval strategy to identify temporally and semantically relevant content from both current and prior visits. This retrieved evidence is used to prompt a large language model (LLM) to generate clinically coherent and temporally aware progress notes.

We evaluate DENSE on a curated cohort of patients with multiple visits and complete progress note documentation. The generated notes demonstrate strong longitudinal fidelity, achieving a temporal alignment ratio of $1.089$, surpassing the continuity observed in original notes. By restoring narrative coherence across fragmented documentation, our system supports improved downstream tasks such as summarization, predictive modeling, and clinical decision support, offering a scalable solution for LLM-driven note synthesis in real-world healthcare settings.
\end{abstract}


\section{Introduction}
\label{sec:introduction}
\IEEEPARstart{C}{linical} documentation lies at the heart of modern healthcare systems. It records patient histories, clinical findings, diagnostic results, interventions, and care decisions throughout a hospital stay. Among these, \emph{progress notes}, written daily by physicians and typically structured using the SOAP format (Subjective, Objective, Assessment, Plan), are essential. They capture a patient's day-to-day evolution in health status and treatment response and serve as a shared reference for care team members \cite{wang2017characterizing,cameron2002soapnotes,rule2021length}. Beyond bedside care, progress notes are foundational to secondary tasks such as clinical summarization, cohort analysis, billing, clinical natural language processing (NLP), and predictive modeling \cite{sterling2019prediction, gao2022hierarchical, gao2025leveraging, kanwal2022attention, lyu2023multimodal}.

However, in real-world electronic health record (EHR) systems, progress notes are often missing, inconsistently documented, or copied forward across days with minimal updates potentially compromising both care quality and downstream data utility \cite{huang2024redundancy, weis2014copy, kuhn2015clinical}. For instance, in the widely used MIMIC-III critical care dataset, only about 8.56\% of hospital visits include any progress notes \cite{johnson2016mimic}. The result is a striking narrative gap: although numerous clinical events take place during hospitalization, there is often no unified document that tells the full story of what transpired each day.

\subsection{Clinical data is fragmented across multiple note types}

Despite the scarcity of progress notes, hospitals collect a wide array of other documentation. A single patient’s visit may include admission notes, radiology reports, nursing assessments, medication reconciliation logs, surgical summaries, consult notes, and discharge planning documents. Each note captures a partial view of care, yet no single note offers a complete narrative.

Table~\ref{tab:notetypes} summarizes the most common note types in MIMIC-III. Collectively, they offer rich and complementary perspectives. For example, a radiology report may describe a chest X-ray finding of pneumonia, a nursing note may mention increased respiratory rate, and a pharmacy note may document antibiotic initiation. Yet, in the absence of a synthesized progress note, reconstructing the full clinical picture is left to human readers or to automated systems often ill-equipped to handle such heterogeneity.

\begin{table*}[ht]
\centering
\caption{Common MIMIC-III clinical note types with descriptions and typical sections.}
\label{tab:notetypes}
\resizebox{\textwidth}{!}{
\begin{tabular}{p{3cm} p{5cm} p{6cm}}
\toprule
\textbf{Note Type} & \textbf{Description} & \textbf{Common Sections} \\
\midrule
Admission note & Documents the patient’s reason for hospitalization and initial exam & Chief Complaint, History of Previous Illness (HPI), Physical Exam \\
Consultation note & Specialist review and recommendations & Reason for Consultation, Assessment, Plan \\
Discharge planning & Coordination and preparation for patient discharge & Education, Home Needs, Follow-up \\
Discharge summary & Final narrative summary of the hospital stay & Hospital Course, Discharge Medications \\
ECG report & Interpretation of electrocardiogram results & Findings, Rhythm Analysis \\
Echo report & Interpretation of echocardiographic imaging & Ejection Fraction, Valve Assessment \\
Event note & Description of significant clinical events & Event Description, Clinical Response \\
Miscellaneous note & Informal or uncategorized documentation & Free-text \\
Nursing other & Observational notes not tied to shift transitions & Fluid Balance, Pain Assessment \\
Nursing shift & Shift-to-shift handoff documentation & Physical Assessment, Interventions \\
Nutrition note & Assessment of dietary needs and intake & Diet, Weight, Recommendations \\
Pharmacy note & Medication history and reconciliation & Medication List, Dose, Route \\
Procedure note & Details about performed clinical procedures & Technique, Complications \\
Progress note & Daily summary of clinical reasoning and patient trajectory & Subjective, Objective, Assessment, Plan \\
Radiology report & Diagnostic interpretation of imaging studies & Impression, Findings \\
Transfer note & Documentation of movement between hospital units & Status, Destination Unit \\
\bottomrule
\end{tabular}
}
\end{table*}

\subsection{Granular challenges in generating progress notes from scattered evidence}

Generating high-quality progress notes from diverse source notes is not a simple extraction task - it requires selection, interpretation, abstraction, and temporal alignment. Several core challenges arise:

\subsubsection{Data sparsity and label inconsistency}  
Few visits in public EHR datasets such as MIMIC-III contain progress notes, limiting supervised training. Even when present, progress notes may be labeled inconsistently (e.g., “Progress note - MICU” vs. “Daily note” vs. “SOAP”), with occasional misspellings like “Dischaarge Plan” or “Nursong Note,” complicating systematic parsing.

\subsubsection{Semantic ambiguity and redundancy}  
Note titles are often ambiguous. A document labeled “Report” may contain radiology findings, nursing assessments, or other content. Similarly, notes titled “DC Plan” and “Discharge Planning” may contain duplicate text, leading to redundancy during input processing \cite{hribar2020redundancy}.

\subsubsection{Lack of temporal linkage across documents}  
Notes are timestamped individually, but lack explicit linkage across a hospital course. This complicates the construction of a coherent chronological timeline especially when timestamps are inconsistent or missing altogether.

\subsubsection{Length variability and computational bottlenecks}  
Note lengths vary dramatically from single-sentence procedure notes to multi-page nursing narratives. This poses challenges for large language models (LLMs) due to input length limits and results in trade-offs between truncation (information loss) and incoherent generation from overly short inputs.

\subsubsection{Clinical reasoning and abstraction}  
Progress notes synthesize diverse findings into clinical assessments (e.g., “Tachycardia likely due to sepsis, given fever and leukocytosis”). This abstraction is a key part of physician reasoning but is difficult for generative models to emulate, especially when source inputs are descriptive rather than inferential.

These challenges highlight the inadequacy of single-note summarization or naïve concatenation approaches. A clinically grounded generation method is required.

\subsection{Problem Statement}

This work addresses the problem of automatically generating structured, clinically relevant, and temporally coherent progress notes for each hospital visit even when no such note exists by synthesizing scattered documentation across heterogeneous EHR inputs.

This task entails:
\begin{itemize}
  \item Normalizing and categorizing noisy real-world notes into consistent formats.
  \item Structuring data across time to form visit-level context windows.
  \item Applying a generation framework that mimics clinical workflows and preserves narrative continuity.
  \item Evaluating output based on both textual fidelity and medical realism.
\end{itemize}

\subsection{Contributions}

To address these challenges, we introduce \textbf{DENSE} (Documenting Evolving Notes from Scattered Evidence), a modular system for longitudinal progress note generation. Our core contributions are:

\begin{enumerate}
    \item \textbf{Unified note taxonomy and temporal alignment:} A clinically informed re-categorization of noisy note labels in MIMIC-III into 16 consistent types (Table~\ref{tab:notetypes}), and a temporal segmentation mechanism that organizes notes into structured visit-centric timelines reflecting clinical workflows.

    \item \textbf{Clinically informed retrieval-augmented generation:} Extending retrieval-augmented generation (RAG) using clinically guided chunking and a vector database to retrieve temporally and semantically relevant evidence. This is then passed, along with prior visit summaries, to a LLM for structured generation. The retrieval approach builds upon the CLI-RAG framework \cite{keerthana2025cli}, adapted here for longitudinal synthesis across note types and visits.

    \item \textbf{Longitudinal evaluation benchmark:} A curated gold-standard set of 56 patients from MIMIC-III, each with 10 to 57 hospital visits and fully documented progress note histories. This allows evaluation of synthetic note trajectories over extended patient timelines.
\end{enumerate}

To our knowledge, this is the first work to (i) temporally align multi-type clinical note data into visit-structured timelines, (ii) generate autoregressive synthetic progress notes using evidence-aware retrieval across visits, and (iii) benchmark note generation in a longitudinal setting.

The remainder of this paper is organized as follows: Section~\ref{sec:related} reviews related work, Section~\ref{sec:data} details dataset curation, Section~\ref{sec:methodology} describes our system architecture, Section~\ref{sec:experiments} presents experimental results, Section~\ref{sec:discussion} outlines implications and future directions, and Section~\ref{sec:conclusion} concludes the paper.

\section{Related Work}
\label{sec:related}

Prior research on clinical note generation spans several domains, including structured-to-text generation, single-visit synthesis, synthetic EHR construction, temporal modeling, and clinical evaluation but few, if any, unify these threads to tackle the synthesis of temporally coherent, visit-level progress notes from heterogeneous clinical evidence~\cite{li2021synthetic}.

Initial efforts focused on translating structured clinical codes (e.g. International Classification of Diseases (ICD), Current Procedural Terminology (CPT) codes into natural language summaries. Melamud et al.~\cite{melamud2019towards} trained neural models to generate patient histories from coded sequences, while Tang et al.~\cite{tang2019progress} emphasized learning patient representations from structured EHR timelines to support clinical classification. However, these approaches typically ignore the unstructured, richly contextualized narratives that physicians write such as progress notes and therefore fall short in replicating the depth of clinical reasoning.

More recently, attention has shifted to using large language models (LLMs) for note synthesis, particularly within single hospital encounters. Soni et al.~\cite{soni2025toward} employed prompt-based generation over structured tabular data to synthesize progress notes, and Biswas et al.~\cite{biswas2024intelligent} applied similar models to simulated provider-patient dialogues. Lu et al.'s ClinicalT5~\cite{lu2022clinicalt5} adapted T5 for clinical summarization tasks by fine-tuning on discharge and radiology note pairs, showing strong results on individual-document generation. However, none of these systems model longitudinal context or integrate information across disparate note types and hospital days. Similarly, Palm et al.~\cite{stetson2012assessing} focused on encounter-level documentation quality using the PDQI-9 rubric, but did not address how notes evolve over time.

Synthetic EHR generators such as Synthea~\cite{walonoski2018synthea}, EMRBots~\cite{kartoun2019advancing}, and medGAN~\cite{choi2017generating} simulate patient-level data for benchmarking and experimentation. These tools are valuable for population-level modeling and data augmentation, yet primarily generate structured outputs like diagnosis codes and encounter sequences. They do not attempt to simulate the narrative, physician-authored documentation that underpins clinical decision-making. Representation models like ClinicalBERT~\cite{huang2019clinicalbert} and Asclepius~\cite{kweon2023publicly} have enabled downstream prediction tasks by improving contextual embeddings of EHR notes, but were not designed for generative tasks.

Longitudinal modeling in clinical NLP has been explored through topic models and dynamic embeddings. For instance, Dynamic Embedded Topic Models (DETM)~\cite{dieng2019dynamic} and Embedded Topic Models (ETM)~\cite{bagheri2020etm} capture evolving disease themes over time by learning topic distributions from patient histories. Such models have been applied to patient trajectory clustering and outcome forecasting~\cite{blei2006dynamic}, but they are not used to generate fluent clinical narratives. In our work, we leverage these thematic frameworks not for generation, but as part of our evaluation: measuring whether the generated progress notes reflect plausible topical shifts over time.

Evaluation of clinical text generation remains an open challenge. Traditional metrics like BLEU and ROUGE offer limited insight into clinical fidelity due to their lexical surface-level comparison~\cite{shool2025systematic}. Palm et al.~\cite{stetson2012assessing} introduced PDQI-9 to assess documentation on clinical dimensions such as completeness, organization, and plausibility, offering a more rigorous alternative. More recently, GPT-based evaluators have been used for realism and coherence assessments, but few works explicitly quantify longitudinal consistency across generated notes. Our evaluation suite extends prior efforts by incorporating semantic similarity, thematic alignment, and GPT-based realism across entire hospitalization trajectories.

Despite this rich body of work, key gaps remain. Existing systems rarely model autoregressive note generation over time, almost never integrate multiple note types (e.g., discharge summaries, consults, and nursing notes), and seldom leverage retrieval-augmented methods tailored to clinical tasks. Furthermore, evaluation remains focused on static text similarity or note-level realism, ignoring how well notes cohere over time.

Our framework, \textbf{DENSE}, bridges these gaps by synthesizing visit-level progress notes that are temporally coherent and clinically grounded. Built on a retrieval-augmented architecture inspired by CLI-RAG~\cite{keerthana2025cli}, DENSE aligns evidence across time and sources, generates notes autoregressively at each visit, and evaluates them for narrative continuity and medical plausibility advancing the frontier of generative modeling in real-world EHR systems.

\section{Data Exploration and Note Categorization}
\label{sec:data}

Electronic Health Records (EHRs) are composed of diverse, free-text clinical notes serving various roles throughout a patient’s care journey. The \textit{NOTEEVENTS} table in the MIMIC-III database~\cite{johnson2016mimic} is a rich source of such documentation, spanning admission notes, radiology reports, ECG interpretations, physician notes, discharge summaries, and nursing observations. These are broadly classified using two metadata fields: \texttt{CATEGORY} and \texttt{DESCRIPTION}, which aim to represent the note type and purpose. However, the structure is highly inconsistent, characterized by extreme lexical variation, semantic ambiguity, and inconsistent labeling, posing challenges for downstream analysis.

Table~\ref{tab:clinical-note-types} outlines the typical clinical significance of the most common note types found in MIMIC-III, informed by the original schema and prior literature

\begin{table}[ht]
\centering
\large
\resizebox{\columnwidth}{!}{
\begin{tabular}{|p{4cm}|p{7.2cm}|}
\hline
\textbf{Note Category} & \textbf{Primary Clinical Content} \\
\hline
Nursing/Other & Generic, unstructured care notes (e.g., ''Report''); may include vitals, observations, shift details. \\
Radiology & Imaging reports (CT, X-ray, MRI), includes anatomical observations, impressions. \\
Discharge Summary & High-level summary of diagnosis, interventions, outcome, and follow-up. \\
Echo / ECG & Specialized test interpretations (cardiac function, electrical activity). \\
Nutrition / Pharmacy & Dietician assessments, medication administration plans, sedation protocols. \\
Physician / Progress Notes & Physician reasoning, status updates, response to treatment, ongoing plan. \\
Consult / Rehab & Specialist evaluations, therapy plans (e.g., PT, OT). \\
General / Social Work / Case Management & Miscellaneous notes-often sparse, ambiguous, or duplicated. \\
\hline
\end{tabular}
}
\caption{Common Note Categories and Their Clinical Relevance in MIMIC-III}
\label{tab:clinical-note-types}
\end{table}

Despite this broad categorization, the actual content and metadata within \texttt{CATEGORY} and \texttt{DESCRIPTION} fields display significant inconsistencies. For instance, under \texttt{CATEGORY='Physician'}, more than 800 distinct \texttt{DESCRIPTION} entries are found, including variants such as ''Physician Resident Progress Note'', ''Progress Note'', ''MICU Resident Progres Note'', and misspellings like ''Physican''. Similarly, the term “Report” appears under multiple categories including \texttt{Nursing/Other}, \texttt{ECG}, \texttt{Radiology}, and \texttt{Echo}, demonstrating both syntactic variability and semantic ambiguity.

A detailed audit of these fields across 2.4 million notes revealed several structural issues:
\begin{enumerate}
    \item \textbf{Syntactic Variability:} Descriptions differ in case, punctuation, and phrasing (e.g., ''progress note'', ''Progress Note'', ''Progress note - MICU'').
    
    \item \textbf{Semantic Ambiguity:} Notes labeled under ''Generic Note'' or ''Report'' span nursing, radiology, and nutrition without meaningful distinction.
    
    \item \textbf{Misspellings and Typos:} Examples include ''Dischaarge Planning Update'', ''Nursong Progress Note'', ''Dishcarge''.
    
    \item \textbf{Sparse Custom Labels:} Long-tail of single-instance note types that defy categorization, such as ''Flumazenil Challenge'', ''Death  Note'', ''Phone call to wife''.
    
    \item \textbf{Ambiguity and Overloading:} Categories such as \texttt{General} and \texttt{Physician} include an eclectic mix of note types including procedure notes, progress notes, and event records. This impedes downstream classification or modeling tasks.
    
    \item \textbf{Redundancy and Noise:} There are numerous near-duplicate or inconsistent labels within \texttt{DESCRIPTION}. For example, “DC Plan”, “Discharge Plan”, and “Dischaarge Planning Update” all refer to the same semantic construct but are logged separately.
    
    \item \textbf{Sparse Documentation:} Only about 8.56\% of all visits in MIMIC-III include progress notes, making it difficult to use them as the primary narrative source for temporal modeling.
    
    \item \textbf{Lack of Temporal Structuring:} The existing schema does not explicitly encode time-aligned sequences of visits or notes across the patient timeline, nor does it link follow-up notes with prior events, despite being essential for longitudinal analysis \cite{zhang2011evaluating, wrenn2010quantifying, liu2022note}.
\end{enumerate}

\begin{figure}[ht]
\centering
\resizebox{0.9\columnwidth}{!}{
\begin{tikzpicture}
\begin{axis}[
    ybar,
    bar width=0.3cm,
    width=\columnwidth,
    height=7cm,
    ylabel={Number of Notes},
    xlabel={Top-Level Category},
    symbolic x coords={Nursing/other,Radiology,Nursing,ECG,Physician,Discharge summary,Echo,Respiratory,Nutrition,General},
    xtick=data,
    x tick label style={rotate=45, anchor=east, font=\footnotesize},
    ymin=0,
    ymajorgrids=true,
    tick label style={font=\footnotesize},
]
\addplot+[blue,fill=blue!30] coordinates {
    (Nursing/other,822497)
    (Radiology,522279)
    (Nursing,223556)
    (ECG,209051)
    (Physician,141624)
    (Discharge summary,59652)
    (Echo,45794)
    (Respiratory,31739)
    (Nutrition,9418)
    (General,8301)
};
\end{axis}
\end{tikzpicture}
}
\caption{Distribution of Notes Across Broad Categories in MIMIC-III (Top 10).}
\label{fig:note_distribution}
\end{figure}
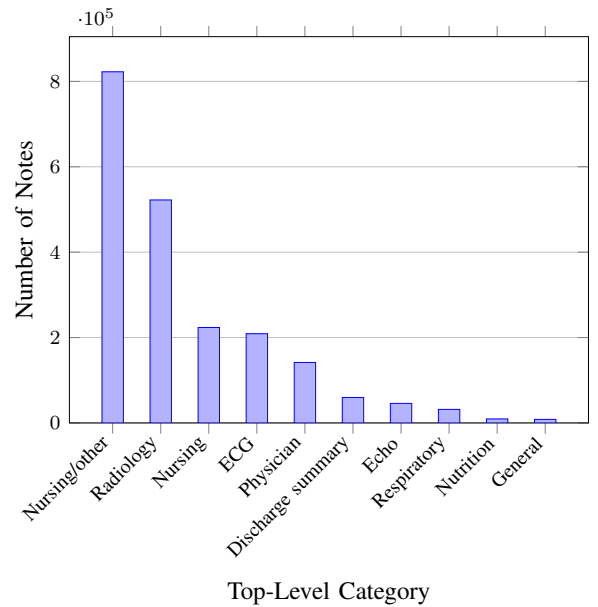

Figure~\ref{fig:note_distribution} shows the distribution of notes across the top 10 broad categories, revealing significant class imbalance. Note: “Nursing/other” refers to miscellaneous nursing documentation (e.g., vital signs, interventions) that does not fall under standardized nursing note templates, whereas “Nursing” includes structured nursing progress notes.

\subsection{Reclassification Strategy}
To resolve these issues, we applied a deterministic re-mapping function that uses regular expressions and semantic pattern matching over cleaned \texttt{DESCRIPTION} fields, with fallback logic using the original \texttt{CATEGORY} when needed. For instance, entries with “progress note” or shift-time patterns were mapped to \texttt{progress\_notes} or \texttt{nursing\_shift\_notes}, respectively, while vague or erroneous labels (e.g., “--error--”) were assigned to \texttt{misc\_notes}. This approach reorganized semantically overlapping or inconsistent labels into 16 coherent and clinically meaningful types (as seen in Table~\ref{tab:note_category_issues}), improving consistency while preserving original coverage.

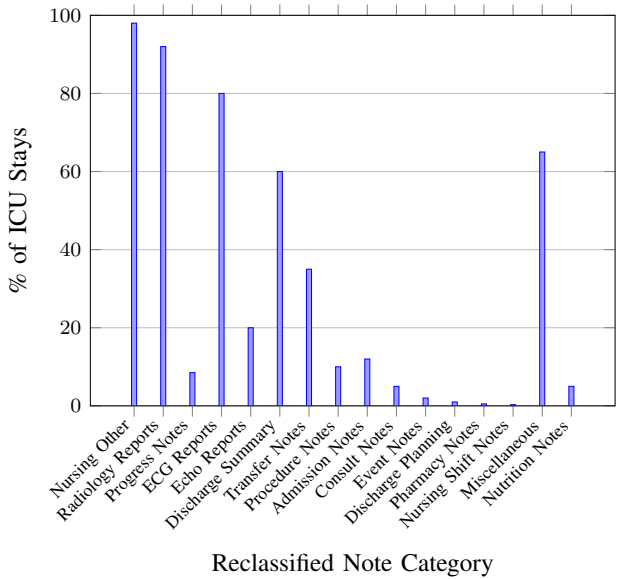
\begin{figure}[ht]
\centering
\resizebox{0.95\columnwidth}{!}{
\begin{tikzpicture}
\begin{axis}[
    ybar,
    bar width=2pt,
    width=\linewidth,
    height=7cm,
    ylabel={\% of ICU Stays},
    xlabel={Reclassified Note Category},
    symbolic x coords={
        Nursing Other,
        Radiology Reports,
        Progress Notes,
        ECG Reports,
        Echo Reports,
        Discharge Summary,
        Transfer Notes,
        Procedure Notes,
        Admission Notes,
        Consult Notes,
        Event Notes,
        Discharge Planning,
        Pharmacy Notes,
        Nursing Shift Notes,
        Miscellaneous,
        Nutrition Notes
    },
    xtick=data,
    x tick label style={rotate=45, anchor=east, font=\scriptsize},
    ymin=0,
    ymax=100,
    ymajorgrids=true,
    tick label style={font=\footnotesize},
]
\addplot+[fill=blue!40] coordinates {
    (Nursing Other, 98)
    (Radiology Reports, 92)
    (Progress Notes, 8.5)
    (ECG Reports, 80)
    (Echo Reports, 20)
    (Discharge Summary, 60)
    (Transfer Notes, 35)
    (Procedure Notes, 10)
    (Admission Notes, 12)
    (Consult Notes, 5)
    (Event Notes, 2)
    (Discharge Planning, 1)
    (Pharmacy Notes, 0.5)
    (Nursing Shift Notes, 0.3)
    (Miscellaneous, 65)
    (Nutrition Notes, 5)
};
\end{axis}
\end{tikzpicture}
}
\caption{Coverage of Each Note Category Across ICU Stays (\% of Patients)}
\label{fig:visit_coverage}
\end{figure}

Figure~\ref{fig:visit_coverage} shows the percentage of ICU stays covered by each reclassified note category. Notably, fewer than 9\% of stays contain true physician-authored progress notes, highlighting a key motivation for synthetic note generation.

\begin{table}[ht]
\caption{Granular Issues in Original MIMIC-III Note Categories and Their Resolution via Semantic Reassignment}
\label{tab:note_category_issues}
\centering
\Large
\resizebox{\columnwidth}{!}{
\begin{tabular}{|p{5cm}|p{5cm}|p{6cm}|}
\toprule
\textbf{Original Issue} & \textbf{Examples} & \textbf{Resolved Category} \\
\midrule
Overloaded ``Nursing/other'' & All 822K entries labeled only ``Report'' & nursing\_other\_notes \\
Sparse or ambiguous ``Physician'' descriptions & ``Progress Note'', ``Attending PN'' & Parsed by regex, e.g. progress notes with time stamps to progress\_notes \\
Multiple ``Discharge Plan'' terms scattered under Case Management & ``DC Plan'', ``Discharge Plan Note'', ``Hospice Referral'' & Mapped to discharge\_planning \\
``Consult'' spread across categories like ``Consult'', ``General'', ``Physician'' & ``Cardiology Consult'', ``GI Consult'', ``Critical Care Consult'' & Consolidated to consult\_notes \\
Ambiguous event types under ``General'' or ``Physician'' & ``Family Meeting'', ``Code Discussion'', ``Death Note'' & Grouped under event\_notes \\
Short notes or error tags & ``--error--'', ``Generic Note'', blank strings & Mapped to fallback category misc\_notes \\
Rich note types hidden in text & e.g. ``thoracentesis'', ``intubation'' in description & Mapped to procedure\_notes \\
\bottomrule
\end{tabular}
}
\end{table}

This refined note taxonomy enables robust downstream applications, including semantic filtering, prompt conditioning, and longitudinal modeling, as described in Section~\ref{sec:methodology}.

\section{Methodology}
\label{sec:methodology}

\begin{figure*}[ht]
\centering
\includegraphics[width=\textwidth]{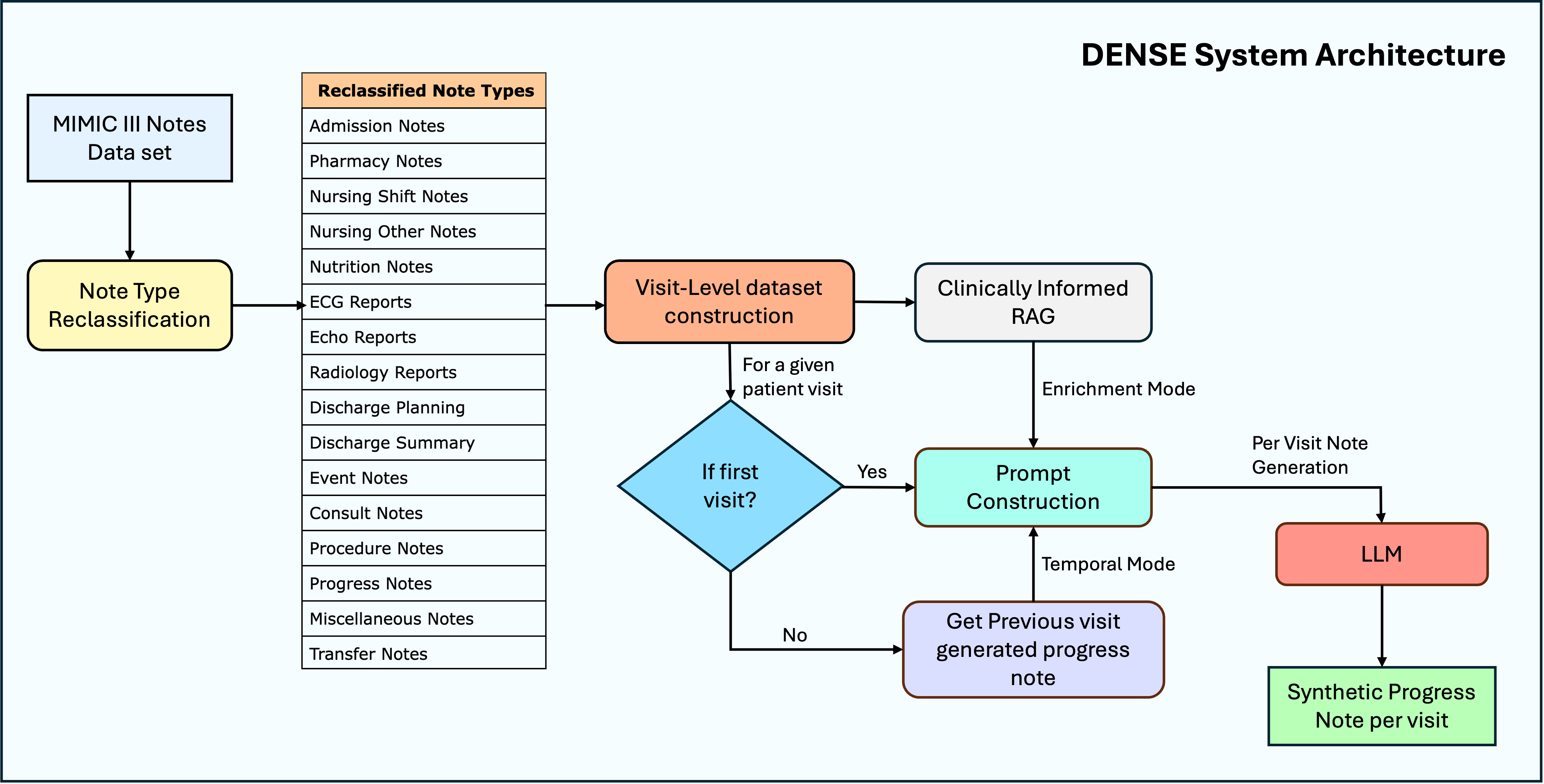}
\caption{DENSE pipeline for generating structured SOAP‑style progress notes. Annotated modules indicate where Clinically informed RAG is integrated as a retrieval component.}
\label{fig:architecture}
\end{figure*}

We present \textbf{DENSE} (Documenting Evolving Progress Notes from Scattered Evidence), a modular pipeline for generating SOAP-style progress notes that are temporally coherent and clinically grounded across patient visits. DENSE combines structured note-type reclassification, visit-level data pivoting, domain-specific preprocessing, hierarchical chunking, semantically filtered retrieval, and LLM-based generation as shown in Figure~\ref{fig:architecture}

\subsection{Data Processing and Visit-Level Construction}

\textbf{Note-Type Reclassification:}  
We use over 2.4 million notes from the \texttt{NOTEEVENTS} table in MIMIC‑III~\cite{johnson2016mimic}. These are inconsistently labeled across broad categories (e.g., ``Physician'', ``Nursing'') and varied subcategories. To resolve this, we apply regular-expression heuristics and manual curation to standardize notes into 16 clinically meaningful types (e.g., \texttt{progress\_notes}, \texttt{radiology\_reports}, \texttt{consult\_notes}, \texttt{nursing\_shift\_notes}). This reclassification mitigates syntactic variability, semantic ambiguity, and category overload, providing a reliable basis for downstream tasks.

\textbf{Visit-Level Pivoting:}  
Using the harmonized note types, we construct a visit-centric dataset. Each row corresponds to a unique hospital admission indexed by \texttt{SUBJECT\_ID} and \texttt{HADM\_ID}, with columns for \texttt{CHARTDATE} and each of the 16 note types containing concatenated content from that visit. This structure supports longitudinal alignment and retrieval across time.

\subsection{Note-Type-Specific Preprocessing}
\textbf{Custom Cleaning Pipelines.}  
Raw clinical notes are highly variable. To normalize content while preserving clinical semantics, we apply note-type-specific preprocessing. This includes standardizing section headers (e.g., \textit{History of Present Illness}, \textit{Assessment and Plan}), removing boilerplate templates, normalizing punctuation and spacing, and converting bulleted/tabular data to natural language.

\textbf{Generic Preprocessing:}  
We apply shared cleaning steps across all notes: Unicode normalization, de-identification (e.g., replacing '[**...**]'), bullet-to-paragraph conversion, and key-value consolidation. However, different note types (e.g., radiology vs. nursing) undergo tailored processing pipelines based on structure and clinical intent. Header templates are defined in a \texttt{NOTE\_HEADERS} dictionary, ensuring section-level integrity for downstream chunking.

\subsection{Evidence Retrieval with CLI-RAG}

DENSE incorporates the CLI-RAG~\cite{keerthana2025cli} retrieval framework to surface semantically relevant evidence for note generation.

\textbf{Hierarchical Chunking:}  
Each preprocessed note is first segmented by clinical headers using curated templates. Long sections are recursively split into overlapping windows (approx. 3000 characters, 300-character overlap). Each chunk is embedded using Sentence-BERT (\texttt{all-MiniLM-L6-v2}) into 384-dimensional vectors and indexed in ChromaDB with metadata (note type, section, visit ID, etc.).

\textbf{Retrieval Strategy:}  
Retrieval operates in two modes:
\begin{itemize}
    \item \textbf{Global retrieval:} Semantic search across all indexed chunks using SOAP-aligned queries (e.g., “What treatments were provided?”).
    \item \textbf{Local retrieval:} Targeted search within specific note types using refined prompts.
\end{itemize}
Retrieved evidence is deduplicated, filtered by recency and section relevance, and chronologically ordered. This step ensures that generated content reflects both thematic and temporal coherence.

\subsection{Progress Note Generation via LLMs}
Once semantically relevant chunks are retrieved for a given hospital visit, we construct structured prompts to guide large language models (LLMs) in generating high-quality, SOAP-style progress notes. These prompts are carefully engineered to integrate multiple sources of information while maintaining clinical fidelity and narrative coherence.

For each visit, the prompting process begins with a preprocessing phase where the retrieved chunks are parsed and summarized to remove redundancy and noise. These chunks originating from up to 16 diverse note types (e.g., radiology, nursing, consults, etc.) are cleaned, section-tagged, and chronologically ordered to reflect the flow of clinical events during the visit. This step produces a synthesized prompt summary capturing the most salient clinical evidence across modalities.

In visits beyond the patient's first, we incorporate a longitudinal aspect by referencing the previously generated progress note from the immediately preceding visit. Instead of simply appending the full earlier note, we produce a concise, clinically meaningful summary of that note, focusing on elements like ongoing problems, treatments, and follow-up plans. This temporal linkage allows the model to simulate the behavior of human clinicians who routinely reference prior notes to maintain continuity of care.

The prompting is carried out in two distinct modes: \textbf{enrichment} and \textbf{temporal}. In the enrichment mode, used exclusively for a patient's first recorded visit, the LLM prompt is composed solely of the semantically relevant evidence chunks gathered via retrieval techniques. These chunks are drawn from whatever subset of the 16 note types is available for that visit and are designed to provide broad, diverse clinical context for note synthesis. In contrast, the temporal mode applied to all subsequent visits augments this enrichment mode prompt with a summary of the previous visit’s progress note, enabling longitudinal reasoning and coherence over time. This dual-mode prompting structure is central to DENSE's ability to replicate how clinicians integrate both current evidence and patient history when documenting care.

\subsection{Summary and Relation to Prior Work}
DENSE advances the state of clinical note generation by integrating longitudinal reasoning into the generation process, something prior systems like CLI-RAG~\cite{keerthana2025cli} do not explicitly support. While CLI-RAG serves as a powerful retrieval backbone for surfacing semantically relevant evidence, its design centers around single-visit summarization. In contrast, DENSE builds on this foundation to support multi-visit narrative construction, modeling how clinicians document a patient's evolving story over time.

This is achieved through several innovations. First, visit-level note aggregation enables the construction of temporally localized document corpora for each care episode. Second, MIMIC-III notes analysis-based preprocessing allow for consistent semantic segmentation of diverse clinical texts. Third, prompt engineering incorporates both enrichment and temporal components, mirroring the dual inputs clinicians rely on current evidence and prior context. Most crucially, by integrating previous synthetic notes into the prompt construction process, DENSE simulates continuity-aware documentation, enabling LLMs to generate progress notes that not only summarize the current visit, but also reflect the broader trajectory of the patient's care.

In doing so, DENSE fills a key gap in the literature by enabling structured, temporally conditioned progress note generation across multiple visits. This capability marks a methodological shift from isolated note synthesis to coherent, longitudinal clinical storytelling.

\section{Experimental Setup and Evaluation}
\label{sec:experiments}
This section details the empirical evaluation of the DENSE framework, focusing on its ability to produce clinically meaningful, temporally consistent progress notes across longitudinal patient trajectories. We describe the cohort selection, evaluation methodology, and provide a comprehensive analysis of results supported by quantitative metrics and visualizations.

\subsection{Dataset and Cohort Selection}
To assess longitudinal note generation, we constructed a cohort of 56 patients from MIMIC-III, each with 10 to 57 documented hospital visits, resulting in a total of over 1,100 encounters. For every visit, DENSE generated a synthetic progress note, simulating what a clinician might document based on available clinical context.

For the first encounter of each patient, only concurrent clinical evidence from various note types was used to simulate documentation from a single point in time. For all subsequent visits, DENSE incorporated both current visit evidence and a summary of the previous visit’s note. This design mimics the natural behavior of clinicians, who reference past documentation to maintain continuity and adapt to a patient’s evolving condition.

The data was structured in a pivoted format where each row represented one visit, indexed by \texttt{SUBJECT\_ID}, \texttt{HADM\_ID}, \texttt{CHARTDATE}, and included 16 standardized clinical note types (e.g., discharge summaries, radiology reports, lab results, etc.) that served as potential sources of contextual information.

\subsection{Evaluation Metrics}
We used a comprehensive suite of evaluation metrics to capture performance across lexical, semantic, structural, and temporal dimensions.

Lexical overlap was measured via BLEU and ROUGE scores. BLEU was predictably low (mean = 0.0116), reflecting DENSE's tendency to paraphrase and reframe information rather than match exact wording. ROUGE-1 (0.2738) and ROUGE-L (0.1102) showed reasonable token and sequence-level similarity, indicating that key medical terms and phrases were preserved.

Semantic similarity was assessed using cosine similarity computed from `all-mpnet-base-v2` sentence embeddings, yielding a strong average score of 0.7398. This suggests that DENSE-generated notes remain faithful to the core meaning of clinician-authored notes.

\begin{figure*}[h]
\centering
\includegraphics[width=\linewidth]{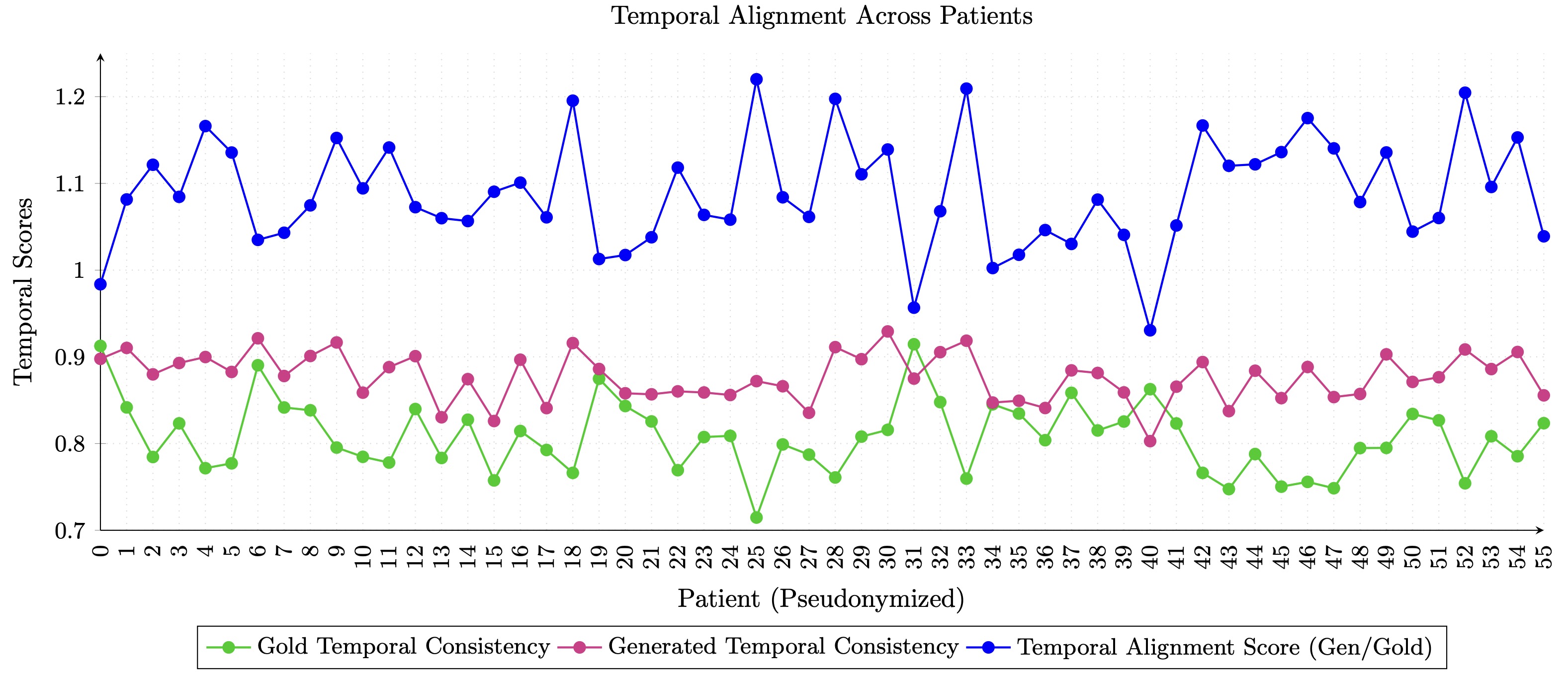}
\caption{Patient-level temporal alignment scores. Higher values reflect better semantic consistency over time. Values above 1.0 indicate higher longitudinal coherence in DENSE-generated notes compared to real notes.}
\label{fig:temporal_scores}
\end{figure*}

SOAP structural completeness was evaluated manually for each note, with all generated notes receiving a perfect 4.0/4.0 score for including the Subjective, Objective, Assessment, and Plan sections.

Length ratio (1.1975) indicated that generated notes were slightly longer on average than gold-standard notes, but remained within a reasonable margin providing additional context without excessive verbosity.

Importantly, to assess longitudinal quality, we introduced a \textbf{temporal alignment metric}. To compute the temporal alignment metric, we measure how semantically coherent a patient's progress notes are across successive hospital visits. Specifically, for each patient, we compute the average \textit{pairwise cosine similarity} between adjacent generated progress notes using embeddings from a Sentence-BERT model (\texttt{all-mpnet-base-v2}). This produces a \textit{temporal consistency score} for the generated notes. We repeat this procedure on the original (gold) notes from MIMIC-III to obtain a baseline consistency score. The final \textit{temporal alignment ratio} is computed as the quotient of the generated score over the gold score:

\begin{equation}
\text{Temporal Alignment Ratio} = \frac{\text{Mean Pairwise Similarity}_{\text{generated}}}{\text{Mean Pairwise Similarity}_{\text{gold}}}
\end{equation}

This ratio captures how well the generated notes maintain longitudinal coherence compared to real clinician-authored notes. A score greater than 1 indicates higher temporal consistency in the synthetic notes. While this is a novel adaptation tailored for our longitudinal setting, it draws conceptually from prior work on semantic similarity and discourse coherence modeling~\cite{barzilay2008modeling, li2014modeling}, but applies it specifically to clinical progress notes over time.

This measured how well adjacent progress notes remained semantically aligned over time, simulating a coherent patient journey. DENSE outperformed the original notes with a mean generated temporal consistency of 0.877 vs. 0.807, yielding an alignment ratio of 1.089 as shown in Table~\ref{tab:results}. Patient-level alignment scores with respective generated temporal consistency and gold temporal consistency is shown in Fig~\ref{fig:temporal_scores}

\begin{table}[ht]
\centering
\caption{Aggregate Evaluation Metrics Across 56 Patients (1,100 Visits)}
\label{tab:results}
\large
\resizebox{\columnwidth}{!}{
\begin{tabular}{|p{5cm}|p{2cm}|p{5cm}|}
\toprule
\textbf{Metric} & \textbf{Mean Score} & \textbf{Interpretation} \\
\midrule
BLEU & 0.0116 & Low n-gram match, expected due to paraphrasing \\
ROUGE-1 & 0.2738 & Reasonable unigram overlap \\
ROUGE-2 & 0.0743 & Limited bigram match, reflects stylistic variation \\
ROUGE-L & 0.1102 & Moderate longest common subsequence \\
Semantic Similarity & 0.7398 & Strong content-level alignment \\
SOAP Structure Score & 4.0 / 4.0 & All sections present in generated notes \\
Length Ratio & 1.1975 & Slightly more verbose, within acceptable range \\
Temporal Consistency (Gold) & 0.807 & Ground truth note consistency \\
Temporal Consistency (Generated) & 0.877 & Better than gold, high longitudinal fidelity \\
Alignment Ratio (Gen/Gold) & 1.089 & Improved narrative stability \\
\bottomrule
\end{tabular}
}
\end{table}

\subsection{Summary}
DENSE demonstrates strong potential for real-world use in longitudinal clinical documentation. Its generated notes maintain semantic fidelity, structural consistency, and most critically temporal continuity, outperforming even gold-standard notes in maintaining coherent clinical narratives across visits. This makes DENSE a valuable foundation for future EHR-integrated generative systems aimed at improving both provider efficiency and care quality.

\section{Discussion}
\label{sec:discussion}

The evaluation confirms that DENSE produces high-fidelity progress notes that are both semantically accurate and temporally coherent. Its design combining structured evidence retrieval with longitudinal conditioning enables continuity across visits, a key challenge in multi-encounter clinical documentation. The strong temporal alignment scores indicate that the system not only captures current clinical context, but also meaningfully integrates prior patient information, improving narrative stability over time.

By framing retrieval around note-type-specific inputs and summarizing past documentation, DENSE shifts generation from surface-level text modeling to grounded clinical synthesis. This behavior is especially valuable for maintaining SOAP structure and preserving clinical intent, even in the presence of paraphrasing and stylistic variation.

\subsection{Clinical and Research Implications}
The ability to generate synthetic, high-quality longitudinal notes has immediate utility in both clinical and research settings - \textbf{Data augmentation} for longitudinal NLP tasks, such as prognosis modeling and phenotyping, \textbf{Filling documentation gaps} in incomplete or fragmented patient records and \textbf{Building intelligent tools} for real-time summarization, scribing, or audit support. These capabilities align with broader efforts to reduce clinician burden, while improving EHR data quality and downstream model performance \cite{palm2025assessing}.

\subsection{Future Directions}
To further advance this work, future efforts should explore integration of clinician feedback for iterative refinement and alignment with documentation standards, extension to MIMIC-IV and real-world EHR systems for external validation, use of generated notes to enhance performance on low-resource or longitudinal prediction tasks and embedding safeguards to detect and mitigate hallucinations in generated content.

\section{Conclusions}
\label{sec:conclusion}

We introduce DENSE, a framework for generating temporally grounded and clinically faithful progress notes by combining structured multi-note retrieval with longitudinal prompt design. Unlike prior methods focused on single-visit summarization, our system incorporates past visit summaries to capture clinical evolution across time, improving coherence and fidelity in multi-visit narratives.

Our evaluation shows that DENSE-generated notes maintain SOAP structure, exhibit strong semantic overlap with reference notes, and outperform ground truth in temporal consistency, demonstrating its utility in modeling continuity of care.

Future work will focus on scaling DENSE to broader patient populations and specialties, integrating multimodal signals such as labs and vitals, and incorporating real-world physician feedback. With robust generalization and potential for downstream utility, DENSE represents a step forward in clinically aligned, AI-assisted documentation.

\section*{References}
\bibliographystyle{IEEEtran}
\bibliography{references}

\end{document}